\documentclass[journal]{IEEEtran}
\usepackage[numbers,sort&compress]{natbib}
\usepackage{amsmath,graphicx}
\usepackage{multirow}
\usepackage{booktabs}
\usepackage{bm}
\usepackage{mathrsfs}
\usepackage{algorithm}
\usepackage{algorithmic}
\usepackage{setspace}
\usepackage{subfigure}
\usepackage[justification=centering]{caption}
\usepackage{float}

\ifCLASSINFOpdf
\else
\fi
\begin{document}
%

\title{\huge {Maximum Correntropy Criterion with Variable Center}}

%
%
%

\author{Badong~Chen,~\IEEEmembership{Senior Member,~IEEE,}
        Xin~Wang,~\IEEEmembership{Student Member,~IEEE,}
        Yingsong~Li,~\IEEEmembership{Member,~IEEE,}
        and~Jose C. Principe,~\IEEEmembership{Member,~IEEE}
\thanks{This work was supported by National Key R\&D Program of China (No. 2017YFB1002501), 973 Program (No. 2015CB351703) and National NSF of China (No. 91648208, No. U1613219).}
\thanks{Badong~Chen and Xin~Wang(chenbd@mail.xjtu.edu.cn and wangxin0420@\newline stu.xjtu.edu.cn) are with the School of Electronic and Information Engineering,Xi'an Jiaotong University, Xi'an 710049, Shaanxi, China.}
\thanks{Yingsong~Li(liyingsong@ieee.org) is with the College of Information and Communication Engineering, Harbin Engineering University, Harbin 150001, China and also with the National Space Science Center, Chinese Academy of Sciences, Beijing 100190, China.}
\thanks{Jose C. Principe(principe@cnel.ufl.edu) is with the School of Electronic and Information Engineering, Xi'an Jiaotong University, Xi'an 710049, Shaanxi, China and also with the Department of Electrical and Computer Engineering, University of Florida, Gainesville, USA.}

}
\maketitle
\begin{abstract}
Correntropy is a local similarity measure defined in kernel space and the maximum correntropy criterion (MCC) has been successfully applied in many areas of signal processing and machine learning in recent years. The kernel function in correntropy is usually restricted to the Gaussian function with center located at zero. However, zero-mean Gaussian function may not be a good choice for many practical applications. In this study, we propose an extended version of correntropy, whose center can locate at any position. Accordingly, we propose a new optimization criterion called maximum correntropy criterion with variable center (MCC-VC). We also propose an efficient approach to optimize the kernel width and center location in MCC-VC. Simulation results of regression with linear in parameters (LIP) models confirm the desirable performance of the new method.
\end{abstract}

\begin{IEEEkeywords}
Correntropy, maximum correntropy criterion (MCC), maximum correntropy criterion with variable center (MCC-VC), robust learning.
\end{IEEEkeywords}

%
\IEEEpeerreviewmaketitle

\section{Introduction}
%
%
%
%
\IEEEPARstart{O}{ne} of the most important problems in machine learning is how to approximate a target random variable ($T$) knowing another ($Y$). This is a central problem in supervised learning, where we design a model ($M$) that receives a random variable $X$ and outputs $Y$ that should approximate $T$ in some sense. The difficulty requires the definition of a loss function (or a similarity measure) to compare $Y$ with $T$. The minimum mean square error (MMSE) criterion is widely used where the loss function is $E\left[ {{e^2}} \right]$, with $e = T - Y$ being the error variable and $E[.]$ the expectation operator. The MMSE is generally computationally simple and mathematically tractable, but its learning performance may degrade seriously when non-Gaussian noises are present in the variables \cite{1principe2010information}.

To improve the learning performance in non-Gaussian noises, a variety of non-MMSE criteria have been proposed in the literature \cite{1principe2010information, 2pei1994least, 3erdogmus2002generalized, 4liu2007correntropy, 5chen2012survival, 6sayin2014novel, 7chen2016generalized, 8chen2016kernel}. Particularly in recent years, the maximum correntropy criterion (MCC) have found many successful applications in domains of signal processing and machine learning, which is very useful for the case where the signals are contaminated by heavy-tailed impulsive noises\cite{9he2011robust,10singh2014c,11hasanbelliu2014information,12ma2015maximum,13feng2015learning,14chen2017maximum,15chen2018mixture}. Under the MCC, an optimal model can be obtained by maximizing the correntropy between the target variable $T$ and the output $Y$ \cite{4liu2007correntropy}:
\begin{equation}
\label{eq1}
\begin{aligned}
{M^*} = \mathop {\arg \max }\limits_{M \in \mathcal{M}} {V_\sigma }(T,Y) = E[{G_\sigma }(e)]
\end{aligned}
\end{equation}
where ${M^*}$ is the optimal model, $\mathcal{M}$ stands for the model¡¯s hypothesis space, and ${V_\sigma }(T,Y) = E[{G_\sigma }({\rm{e}})]$ denotes the correntropy between $T$ and $Y$, with ${G_\sigma }(e)$ being the Gaussian kernel function:
\begin{equation}
\label{eq2}
\begin{aligned}
{G_\sigma }(e) = \frac{1}{{\sqrt {2\pi } \sigma }}exp\left( { - \frac{{{e^2}}}{{2{\sigma ^2}}}} \right)
\end{aligned}
\end{equation}
where $\sigma$ is the kernel bandwidth. Since the Gaussian kernel function ${G_\sigma }(e)$ is a local function of the error variable $e$, the correntropy can be used as an outlier-robust error measure in signal processing and machine learning \cite{1principe2010information}. However, the center of the Gaussian kernel in correntropy is always located at zero, which may not be a good choice for many practical situations. In particular, when the error distribution is non-zero-mean, the original correntropy may perform poorly, because in this case the zero-mean Gaussian function usually cannot match well the error distribution. The goal of the present paper is thus to extend the correntropy to the case where the center can be located anywhere, which potentially can significantly improve the learning performance but is still not fully appreciated in the community.

The rest of the paper is organized as follows. In section II, we define the correntropy with variable center and propose the maximum correntropy criterion with variable center (MCC-VC). In section III, we propose an efficient approach to optimize the kernel width and center location in MCC-VC. Simulation results of regression with linear in parameters (LIP) models are then presented in section IV. Finally, conclusion is given in section V.
\section{Maximum Correntropy Criterion with Variable Center}

In this work, we define the correntropy with variable center between $T$ and $Y$ as follows:
\begin{small}
\begin{equation}
\label{eq3}
\begin{aligned}
{V_{\sigma ,c}}(T,Y) = E[{G_\sigma }(e - c)] = E[\frac{1}{{\sqrt {2\pi } \sigma }}exp\left( { - \frac{{{{\left( {e - c} \right)}^2}}}{{2{\sigma ^2}}}} \right)]
\end{aligned}
\end{equation}
\end{small}
where $c \in \rm{\textbf{R}}$ is the center location. The above definition will reduce to the original correntropy ${V_\sigma }(T,Y)$ when $c = 0$.

Similar to the original correntropy \cite{4liu2007correntropy}, the correntropy with center $c$ also involves all the even moments of the error $e = T - Y$ about the center $c$, that is
\begin{equation}
\label{eq4}
\begin{aligned}
{V_{\sigma ,c}}(T,Y) = \frac{1}{{\sqrt {2\pi } \sigma }}\sum\limits_{n = 0}^\infty  {\frac{{{{( - 1)}^n}}}{{{2^n}n!}}E\left[ {\frac{{{{(e - c)}^{2n}}}}{{{\sigma ^{2n}}}}} \right]}
\end{aligned}
\end{equation}
As $\sigma$ increases, the high-order moments about the center $c$ will decay faster, hence the second-order moment tends to dominate the value. Particularly, when $c = E[e]$ and $\sigma  \to \infty$ , maximizing the correntropy with center $c$ will be equivalent to minimizing the error's variance.

In addition, when the Gaussian kernel shrinks to zero ($\sigma  \to 0$), the correntropy with center $c$ approaches the value of $\int_{ - \infty }^\infty  {{p_{TY}}(t,t - c)dt}$, where ${p_{TY}}(t,y)$ is the joint probability density function (PDF) of $(T,Y)$. This can easily be proved as follows£º
\begin{equation}
\label{eq5}
\begin{aligned}
\mathop {\lim }\limits_{\sigma  \to 0} {V_{\sigma ,c}}(T,Y)&= \mathop {\lim }\limits_{\sigma  \to 0}\iint{G_\sigma }(t - y - c){p_{TY}}(t,y)dtdy \\
&= \iint{\delta }(t - y - c){p_{TY}}(t,y)dtdy\\
&= \int_{ - \infty }^\infty  {{p_{TY}}(t,t - c)dt}
\end{aligned}
\end{equation}
where $\delta (.)$ denotes the Dirac delta function. In this case, we also have
\begin{equation}
\label{eq6}
\begin{aligned}
\mathop {\lim }\limits_{\sigma  \to 0} {V_{\sigma ,c}}(T,Y) &= \mathop {\lim }\limits_{\sigma  \to 0} \int {{G_\sigma }(\varepsilon  - c){p_e}(\varepsilon )d\varepsilon } \\
&= \int {\delta (\varepsilon  - c){p_e}(\varepsilon )d\varepsilon } \\
&= {p_e}(c)
\end{aligned}
\end{equation}

Therefore, when $\sigma  \to 0$, the correntropy with center $c$ will also approach the value of ${p_e}(\varepsilon )$ evaluated at $\varepsilon  = c$, where ${p_e}(.)$ denotes the error's PDF.

The optimal model under the maximum correntropy criterion with variable center (MCC-VC) is defined by
\begin{equation}
\label{eq7}
\begin{aligned}
{M^*} = \mathop {\arg \max }\limits_{M \in \mathcal{M}} {V_{\sigma ,c}}(T,Y) = E[{G_\sigma }(e - c)]
\end{aligned}
\end{equation}

To demonstrate how to solve the optimal solution with finite training samples (by optimizing an empirical risk function), we consider the following linear in parameter (LIP) model:
\begin{equation}
\label{eq8}
\begin{aligned}
{y_i} &= \bm{{h_i}\beta} = \left[ {{\phi _1}({\bm{x}_i}),{\phi _2}({\bm{x}_i}), \cdots ,{\phi _{\tilde N}}({\bm{x}_i})} \right]{\left[ {{\beta _1},{\beta _2}, \cdots ,{\beta _{\tilde N}}} \right]^T}\\
,i &= 1,2, \cdots N
\end{aligned}
\end{equation}
where $\{ {\bm{x}_i},{y_i}\} _{i = 1}^N$ are the $N$ input-output samples, ${\bm{h_i}} = \left[ {{\phi _1}({\bm{x}_i}),{\phi _2}({\bm{x}_i}), \cdots ,{\phi _{\tilde N}}({\bm{x}_i})} \right] \in {\textbf{R}^{\tilde N}}$ is the $i$-th nonlinearly mapped input vector (a row vector), with ${\phi _j}(.)$ being the $j$-th  nonlinear mapping function $(j = 1,2, \cdots \tilde N)$, and $\bm{\beta}  = {\left[ {{\beta _1},{\beta _2}, \cdots ,{\beta _{\tilde N}}} \right]^T} \in {\textbf{R}^{\tilde N}}$ is the output weight vector that needs to be learned. Given $N$ target samples $\{ {t_i}\} _{i = 1}^N$, the output weight vector $\bm{\beta}$ can be trained by minimizing the following regularized MMSE cost:
\begin{equation}
\label{eq9}
\begin{aligned}
{J_{MMSE}}(\beta ) = {\left\| {\textbf{T} - \textbf{Y}} \right\|^2} + \lambda {\left\| \bm{\beta}  \right\|^2}
\end{aligned}
\end{equation}
where $\textbf{Y} = {\left[ {{y_1},{y_2}, \cdots ,{y_N}} \right]^T}$ , $\textbf{T} = {\left[ {{t_1},{t_2}, \cdots ,{t_N}} \right]^T}$, and $\lambda  \ge 0$ is the regularization parameter. In this case, the optimal solution can easily be obtained as
\begin{equation}
\label{eq10}
\begin{aligned}
{\bm{\beta} ^{\rm{*}}} = {\left( {{\textbf{H}^T}\textbf{H} + \lambda \textbf{I}} \right)^{ - 1}}{\textbf{H}^T}\textbf{T}
\end{aligned}
\end{equation}
where $ \rm{\textbf{H}}= \left[ {{\emph{h}_{\emph{ij}}}} \right]$ is an $N \times \tilde N$ dimensional matrix with ${h_{ij}} = {\phi _j}({\bm{x}_i})$. Similarly, one can solve $\bm{\beta}$ by minimizing the following regularized MCC-VC cost:
\begin{equation}
\label{eq11}
\begin{aligned}
{J_{MCC{\rm{ - }}VC}}(\bm{\beta} ) =  - \frac{1}{N}\sum\limits_{i = 1}^N {\left[ {{G_\sigma }({e_i} - c)} \right]}  + \lambda {\left\| \bm{\beta}  \right\|^2}
\end{aligned}
\end{equation}
where ${e_i} = {t_i} - {y_i} = {t_i} - \bm{{h_i}\beta}$ is the $i$-th error sample. Setting $\frac{\partial }{{\partial \bm{\beta} }}{J_{MCC{\rm{ - }}VC}}(\bm{\beta} ) = 0$, one can derive
\begin{equation}
\label{eq12}
\begin{aligned}
{\bm{\beta} ^{\rm{*}}} = {[{{\bf{H}}^T}{\bf{\Lambda H}} + \lambda '{\bf{I}}]^{ - 1}}{{\bf{H}}^T}{\bf{\Lambda }}\textbf{T}'
\end{aligned}
\end{equation}
where $\lambda ' = 2N\lambda$, $\textbf{\rm{T}}' = {[{t_1} - c,{t_2} - c, \ldots ,{t_N} - c]^T}$, and ${\bf{\Lambda }}$ is a diagonal matrix with diagonal elements ${{\bf{\Lambda }}_{ii}} = {G_\sigma }({e_i} - c)$.

The solution (12) is a fixed-point equation since the diagonal matrix ${\bf{\Lambda }}$ on the right-hand side depends on the weight vector $\bm{\beta}$ via ${e_i} = {t_i} - \bm{{h_i}\beta} $. Therefore, the optimal solution under MCC-VC can be solved by using the following fixed-point iteration:
\begin{equation}
\label{eq13}
\begin{aligned}
{\bm{\beta} _k} = {\left. {\left( {{{[{{\bf{H}}^T}{\bf{\Lambda H}} + \lambda '{\bf{I}}]}^{ - 1}}{{\bf{H}}^T}{\bf{\Lambda }}\bf{T}'} \right)} \right|_{{\bm{\beta} _{k - 1}}}}
\end{aligned}
\end{equation}
where ${\bm{\beta} _k}$ is the estimated weight vector at the $k$-th iteration.
\section{Optimization of the Free Parameters in MCC-VC}

There are two free parameters in MCC-VC, namely the kernel width $\sigma$ and the center location $c$, whose values have significant influence on the learning performance. In this section, we propose an efficient approach to optimize the two parameters. First, we divide the correntropy with center $c$ into three terms:
\begin{equation}
\label{eq14}
\begin{aligned}
{V_{\sigma ,c}}(T,Y) &= \int {{G_\sigma }(\varepsilon  - c){p_e}(\varepsilon )d\varepsilon } \\
&= \frac{1}{2}\int {{{\left[ {{G_\sigma }(\varepsilon  - c)} \right]}^2}d\varepsilon }  + \frac{1}{2}\int {{{\left[ {{p_e}(\varepsilon )} \right]}^2}d\varepsilon }  \\&- \frac{1}{2}\int {{{\left[ {{G_\sigma }(\varepsilon  - c) - {p_e}(\varepsilon )} \right]}^2}d\varepsilon }
\end{aligned}
\end{equation}

Since the first term is independent of the model $M$, we have
\begin{equation}
\label{eq15}
{M^*} = \mathop {\arg \max }\limits_{M \in \mathcal{M}} {V_{\sigma ,c}}(T,Y) = \mathop {\arg \max }\limits_{M \in \mathcal{M}} {U_{\sigma ,c}}(T,Y)
\end{equation}
where ${U_{\sigma ,c}}(T,Y) = \int {{{\left[ {{p_e}(\varepsilon )} \right]}^2}d\varepsilon }  - \int {{{\left[ {{G_\sigma }(\varepsilon  - c) - {p_e}(\varepsilon )} \right]}^2}d\varepsilon } $. Then we propose the following optimization:
\begin{equation}
\label{eq16}
\left( {{M^*},{\sigma ^*},{c^*}} \right) = \mathop {\arg \max }\limits_{M \in \mathcal{M},\sigma  \in \mathcal{S},c \in \mathcal{C}} {U_{\sigma ,c}}(T,Y)
\end{equation}
where $\mathcal{S}$ and $\mathcal{C}$ denote the admissible sets of parameters $\sigma$ and $c$. Thus, the model $M$, the kernel width $\sigma$ and the center location $c$ are jointly optimized to maximize the function ${U_{\sigma ,c}}(T,Y)$. To simplify the optimization, we adopt an alternative optimization approach:

i) When the model is fixed(hence the error's distribution is fixed), the term $\int {{{\left( {{p_e}(\varepsilon )} \right)}^2}d\varepsilon } $ is independent of $\sigma$ and $c$, in this case the two free parameters can simply be optimized by
\begin{equation}
\label{eq17}
\begin{aligned}
\left( {{\sigma ^*},{c^*}} \right) &= \mathop {\arg \min }\limits_{\sigma  \in \mathcal{S},c \in \mathcal{C}} \int {{{\left[ {{G_\sigma }(\varepsilon  - c) - {p_e}(\varepsilon )} \right]}^2}d\varepsilon } \\&= \mathop {\arg \min }\limits_{\sigma  \in \mathcal{S},c \in \mathcal{C}} \left\{ {\int {{{\left[ {{G_\sigma }(\varepsilon  - c)} \right]}^2}d\varepsilon }  - 2E\left[ {{G_\sigma }(e - c)} \right]} \right\}\\
&= \mathop {\arg \min }\limits_{\sigma  \in \mathcal{S},c \in \mathcal{C}} \left\{ {\frac{1}{{2\sqrt \pi  \sigma }} - 2E\left[ {{G_\sigma }(e - c)} \right]} \right\}
\end{aligned}
\end{equation}

ii) After the parameters have been determined, the model $M$ can then be optimized by maximizing the function (16) or (14) with $\sigma=\sigma ^*$ and $c=c^*$.

The above procedure can be repeated until convergence.

From (17), one can see that the parameters $\sigma$ and $c$ are optimized such that the Gaussian kernel function ${G_\sigma }(\varepsilon  - c)$ matches the error's PDF ${p_e}(\varepsilon )$ as closely as possible. This is in principle consistent with our intuition. The idea of PDF matching has been explored with great success in the literature of information theoretic learning (ITL) \cite{1principe2010information,16erdogmus2002error,17santamaria2002adaptive,heravi2018new}.
Given $N$ error samples $\{ {e_i}\} _{i = 1}^N$, we have $E\left[ {{G_\sigma }(e - c)} \right] \approx \frac{1}{N}\sum\limits_{i = 1}^N {{G_\sigma }({e_i} - c)}$. It follows that
\begin{equation}
\label{eq18}
\begin{aligned}
\left( {{\sigma ^*},{c^*}} \right) = \mathop {\arg \min }\limits_{\sigma  \in \mathcal{S},c \in \mathcal{C}} \left\{ {\frac{1}{{2\sqrt \pi  \sigma }} - \frac{2}{N}\sum\limits_{i = 1}^N {{G_\sigma }({e_i} - c)} } \right\}
\end{aligned}
\end{equation}

Remark: There are several approaches to solve the optimization problem in (18). For example, we can use a gradient based method to search the solution. In many practical situations, we often find the optimal solution in a given finite set. To further simplify the computation, one can just set the parameter $c$ to the mean or median value of the error samples, and only optimize the kernel width $\sigma$.

Based on the above parameters optimization strategy, a robust regression algorithm with LIP models under MCC-VC can be obtained, which is referred to as the LIP-MCC-VC and is described in Algorithm 1.

\begin{algorithm}\footnotesize
	\renewcommand{\algorithmicrequire}{\textbf{Input:}}
	\renewcommand{\algorithmicensure}{\textbf{Output:}}
	\caption{LIP-MCC-VC}
	\label{alg:1}
	\begin{algorithmic}[1]
		\REQUIRE training samples $\{ {\bm{x}_i},{t_i}\} _{i = 1}^N$, number of nonlinear mappers $\tilde N$, regularization parameter $\lambda '$, maximum iteration number $K$, a set of kernel widths $\mathcal{S}$, a set of kernel centers $\mathcal{C}$, termination tolerance $\xi$ and the initial weight vector ${\bm{\beta} _0}{\rm{ = }}\textbf{0}$.
		\ENSURE weight vector $\bm{\beta}$\\
		\FORALL{$k = 1,2,...,K$}
		\STATE Compute the errors based on ${\beta _{k - 1}}$: ${e_i} = {t_i} - {\bm{h}_i}{\bm{\beta} _{k - 1}}$, $i = 1,2, \cdots ,N$
		\STATE Optimize the parameters $\sigma$ and $c$: $\left( {{\sigma ^*},{c^*}} \right) = \mathop {\arg \min }\limits_{\sigma  \in \mathcal{S},c \in \mathcal{C}} \left\{ {\frac{1}{{2\sqrt \pi  \sigma }} - \frac{2}{N}\sum\limits_{i = 1}^N {{G_\sigma }({e_i} - c)} } \right\}$
		\STATE Compute the diagonal matrix ${\bf{\Lambda }}$: ${\bm{\Lambda} _{ii}} = {G_{{\sigma ^*}}}({e_i} - {c^*})$, $i = 1,2, \cdots ,N$
		\STATE Update the weight vector $\bm{\beta}$: ${\bm{\beta} _k} = {\left. {\left( {{{[{{\bf{H}}^T}{\bf{\Lambda H}} + \lambda '{\bf{I}}]}^{ - 1}}{{\bf{H}}^T}{\bf{\Lambda }}\bf{T}'} \right)} \right|_{{\bm{\beta} _{k - 1}}}}$\\
        \STATE \textbf{Until} $\left| {{J_{MCC{\rm{ - }}VC}}({\bm{\beta} _k}) - {J_{MCC - VC}}({\bm{\beta} _{k - 1}})} \right| < \xi $
		\ENDFOR
	\end{algorithmic}
\end{algorithm}

\section{Simulation Results}

In this section, we present simulation results of regression with LIP models to demonstrate the performance of the proposed method. We consider two LIP models, one is the linear regression model and the another is the extreme learning machine (ELM) \cite{18huang2006extreme,19miche2010op,20huang2012extreme,21xing2013training}, a kind of single hidden layer feed forward neural network (SLFN), in which the input weights and biases of the hidden layer are randomly generated, and only the weights of the output layer need to be trained.

\subsection{Linear Regression}

Consider a simple example in which the data are generated by a two-dimensional linear system ${y_i} = {\bm{w}^*}^T{\bm{x}_i} + {\rho _i}$, where ${\bm{w}^*} = {[1,2]^T}$ and ${\rho _i}$ is an additive noise. The input samples $\{ {\bm{x}_i}\}$ are uniformly distributed over $[ - 2,2] \times [ - 2,2]$. The noise ${\rho _i}$ comprises two mutually independent noises, namely the inner noise ${B _i}$ and the outlier noise ${O_i}$. Specifically, ${\rho _i}$ is given by $\rho {}_i = {\rm{(}}1 - {g_i}{\rm{)}}{B_i} + {g_i}{O_i}$, where ${g _i}$ is a binary variable with probability mass $\Pr {\rm{\{ }}{g_i} = 1{\rm{\} }} = p$, $\Pr {\rm{\{ }}{g_i} = 0{\rm{\} }} = 1 - p$, $(0 \le p \le 1)$, which is assumed to be independent of both $B_i$ and $O_i$. In this example, $p$ is set at $0.1$, and the outlier $O_i$ is drawn from a zero-mean Gaussian distribution with variance $10000$. For the inner noise $B_i$, we consider four zero-mean or non-zero-mean distributions: 1) $\mathcal{N}$(0,2), where $\mathcal{N}(u,\sigma^2)$ denotes the Gaussian PDF with mean $u$ and variance ${\sigma ^2}$; 2) $\mathcal{N}$(3,1); 3) Laplace distribution with zero-mean and variance 1; 4) Chi-square distribution with three degrees of freedom. The root mean squared error (RMSE) is employed to measure the performance, computed by $RMSE = \sqrt {\frac{1}{2}{{\left\| {{\bm{w}_k} - {\bm{w}^*}} \right\|}^2}}$, where $\bm{w}_k$ and $\bm{w}^*$ denote the estimated and the target weight vectors respectively.

We compare the performance of three optimization criteria, namely MMSE, MCC and MCC-VC. For MMSE, there is a closed-form solution, so no iteration is needed. For MCC and MCC-VC, a fixed-point iteration is used to solve the model (see \cite{22chen2015convergence} for the fixed-point algorithm under MCC). The mean $\pm$ deviation results of the RMSE and the training time averaged over 100 Monte Carlo runs are presented in Table I. In the simulation, the sample number is $N$ = 400, the iteration number is $K$ = 100, and the initial weight vector is set to ${\bm{w}_0} = {[{\rm{0}},{\rm{0}}]^T}$. For each criterion, the parameters are selected by trial-and-error to achieve the best results, except that the kernel bandwidth and center location of MCC-VC are chosen through solving the optimization (18). The finite kernel bandwidth set $\mathcal{S}$ is equally spaced over $[0.2,5.0]$ with step size 0.2, and the center set $\mathcal{C}$ is equally spaced over $[-5.0,5.0]$ with step size 0.1. From Table I, we observe: i) MCC and MCC-VC can significantly outperform MMSE although both have no closed-form solution; ii) MCC-VC can achieve better performance than MCC especially for non-zero-mean noises because the cost function center can be set at proper value according to the error PDF adaptively; iii) MCC-VC can save much time through solving (18) to find the best values of parameters $\sigma$ and $c$, without performing trial-and-error to optimize the two parameters. Under the noise of case 2), the error distribution and corresponding Gaussian kernel function $G_{\sigma^*}(e - c^*)$ optimized by (18) at the first and second fixed-point iterations of MCC-VC are shown in Fig. 1. As expected,  the Gaussian kernel function matches the error distribution very well.

\begin{table*}[!htbp]\footnotesize
\centering
\caption{RMSE AND COMPUTING TIME (sec) OF DIFFERENT CRITERIA}\label{RMSE}
\begin{tabular}{ccccc}
  \hline
  & &MMSE&MCC&MCC-VC \\
  \hline
  \multirow{2}{*}{case 1)}&RMSE& $1.2374\pm0.6840$  &\textbf{0.0765}$\pm\textbf{0.0422}$  &$0.0902\pm0.0547$\\
  &TIME(sec)&N/A&$1.2217\pm0.0269$ &$\textbf{0.0962}\pm\textbf{0.0023}$\\
  \multirow{2}{*}{case 2)}&RMSE& $1.2214\pm0.6441$  &$0.1375\pm0.0737$  &$\textbf{0.0505}\pm\textbf{0.0272}$\\
  &TIME(sec)&N/A&$1.2214\pm0.0253$ &$\textbf{0.0976}\pm\textbf{0.0024}$\\
  \multirow{2}{*}{case 3)}&RMSE& $1.2435\pm0.6218$  &$0.0337\pm0.0168$  &$\textbf{0.0332}\pm\textbf{0.0165}$\\
  &TIME(sec)&N/A&$1.2805\pm0.0613$ &$\textbf{0.0957}\pm\textbf{0.0032}$\\
  \multirow{2}{*}{case 4)}&RMSE& $1.1317\pm0.5763$  &$0.1546\pm0.0762$  &$\textbf{0.0910}\pm\textbf{0.0441}$\\
  &TIME(sec)&N/A&$1.2157\pm0.0249$ &$\textbf{0.0978}\pm\textbf{0.0022}$\\
  \hline
\end{tabular}
\end{table*}

\begin{figure*}[htbp]
\setlength{\abovecaptionskip}{0pt}
\setlength{\belowcaptionskip}{0pt}
\centering
\subfigure[]{
\includegraphics[width=2.5in,height=1.9in]{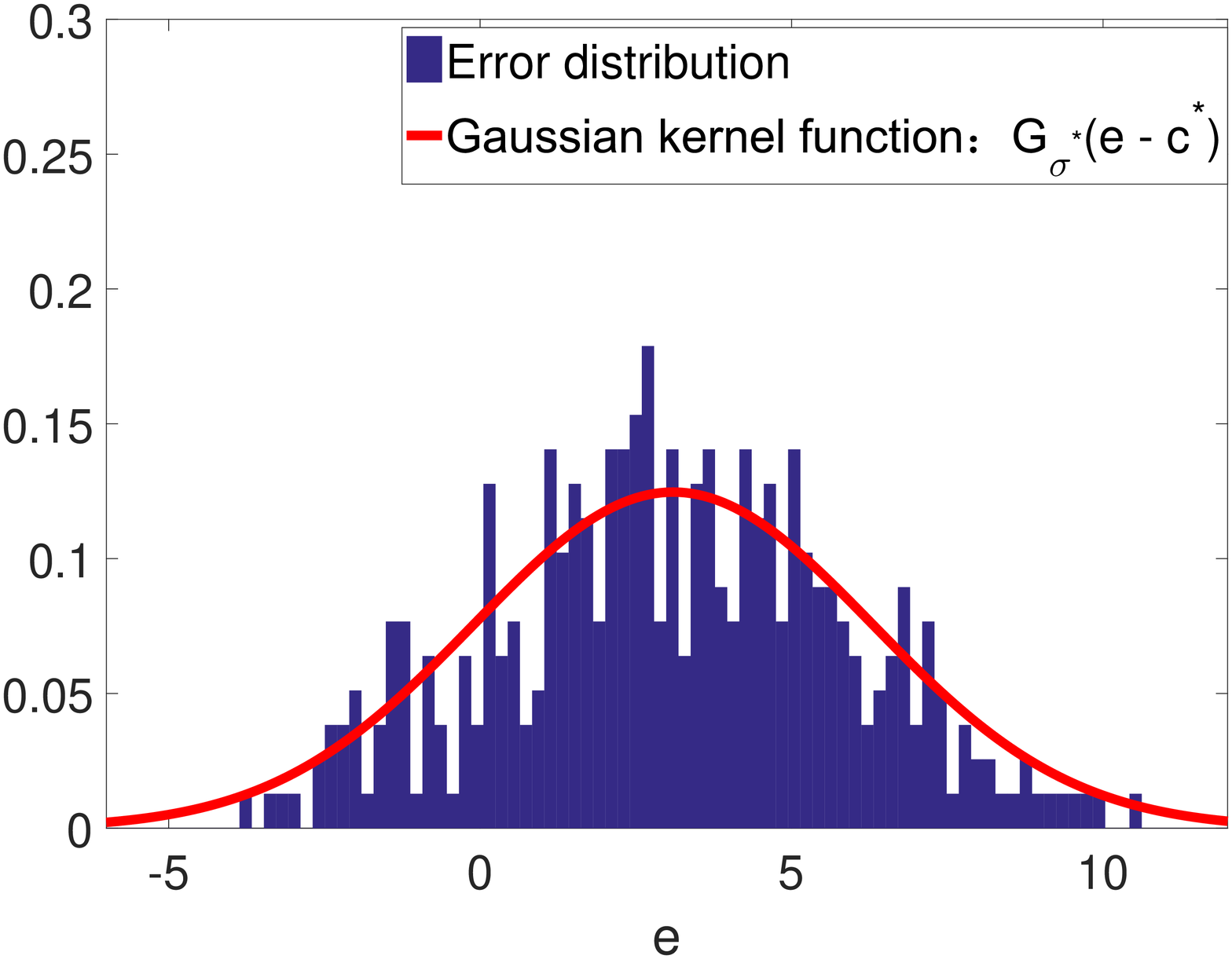}}
\subfigure[]{
\includegraphics[width=2.5in,height=1.9in]{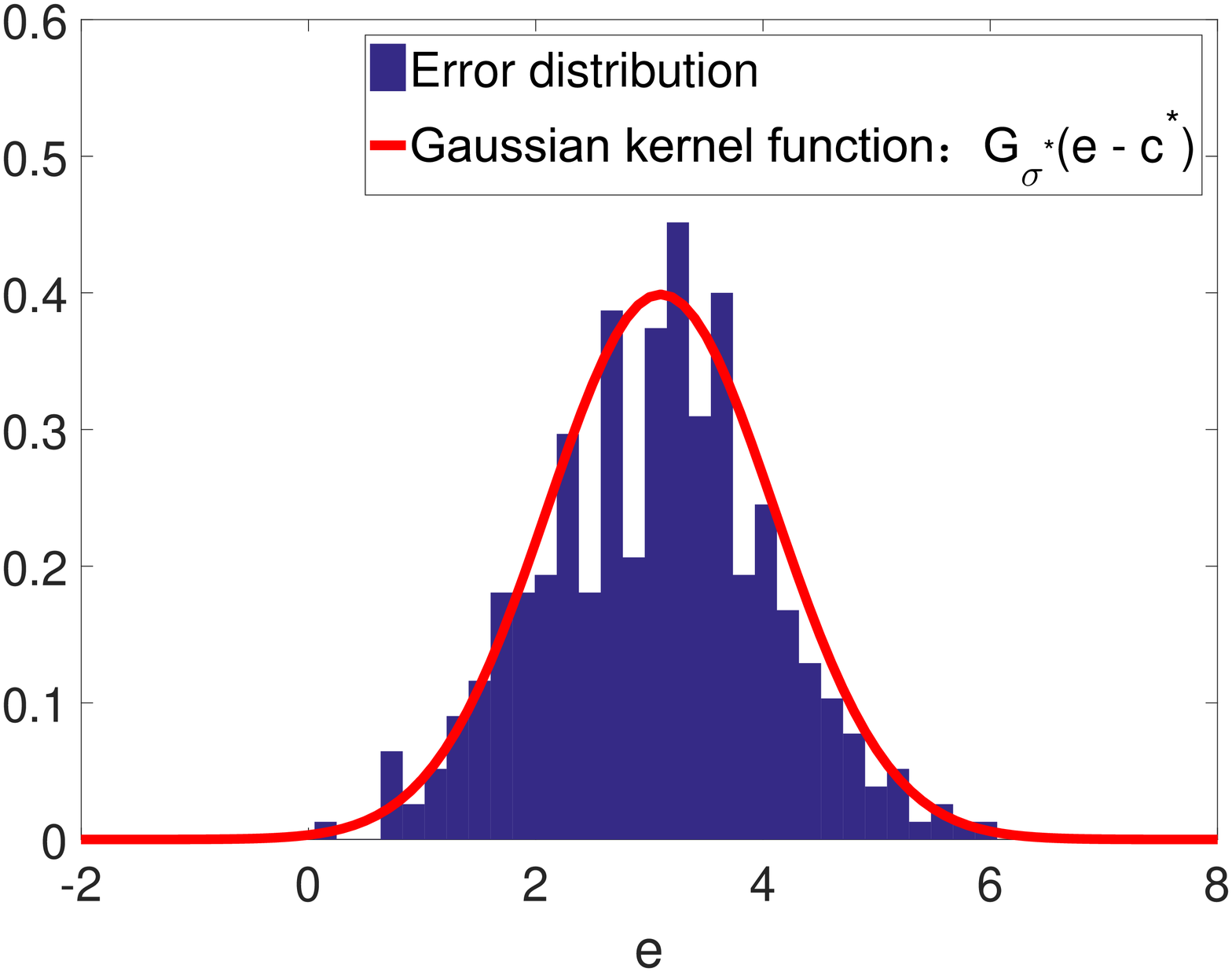}}
\caption{Error distribution and Gaussian kernel function: (a)First iteration, (b)Second iteration.}
\label{fig1}
\end{figure*}
\begin{table*}[]\footnotesize
	\renewcommand\arraystretch{1.5}
	\setlength{\abovecaptionskip}{0pt}
	\setlength{\belowcaptionskip}{5pt}
	\centering
	\caption{ TRAINING AND TESTING RMSEs OF THREE ALGORITHMS}
	\begin{tabular}{p{1.5cm}p{2.5cm}p{2.5cm}p{2.5cm}p{2.5cm}p{2.5cm}p{2.5cm}}
		\toprule
		\multicolumn{1}{c}{\multirow{2}*{Datasets}} & \multicolumn{2}{c}{RELM}& \multicolumn{2}{c}{ELM-RCC}& \multicolumn{2}{c}{ELM-MCC-VC}\\
		\cline{2-7}
		&\multicolumn{1}{c}{Training RMSE} &\multicolumn{1}{c}{Testing RMSE}&\multicolumn{1}{c}{Training RMSE} &\multicolumn{1}{c}{Testing RMSE} &\multicolumn{1}{c}{Training RMSE} &\multicolumn{1}{c}{Testing RMSE}\\
        Servo&$0.0600\pm0.0095$&$ 0.1088\pm0.0171$&$ 0.0831\pm0.0219$&$0.1064\pm0.0165$&$0.0835\pm0.0225$&$\textbf{0.1029}\pm\textbf{0.0179}$\\
        Airfoil&$ 0.0974\pm0.0074$&$0.1031\pm0.0077$&$ 0.0942\pm0.0022$&$0.0997\pm0.0028$&$0.0812\pm0.0038$&$\textbf{0.0923}\pm\textbf{0.0054}$\\
        Concrete&$0.0738\pm0.0021$&$ 0.0965\pm0.0055$&$0.0823\pm0.0025$&$0.0945\pm0.0034$&$0.0642\pm0.0033$&$\textbf{0.0927}\pm\textbf{0.0049}$\\
        Housing&$0.0439\pm0.0042$&$ 0.0921\pm0.0137$&$0.0442\pm0.0042$&$0.0907\pm0.0138$&$0.0455\pm0.0040$&$\textbf{0.0903}\pm\textbf{0.0137}$\\
        Yacht&$ 0.0366\pm0.0093$&$0.0823\pm0.0090$&$0.0575\pm0.0023$&$0.0769\pm0.0053$&$0.0041\pm0.0003$&$\textbf{0.0232}\pm\textbf{0.0105}$\\
        Wine-red&$0.1205\pm0.0036$&$0.1350\pm0.0044$&$0.1171\pm0.0027$&$0.1309\pm0.0035$&$0.1209\pm0.0025$&$\textbf{0.1299}\pm\textbf{0.0031}$\\
        Slump&$0.0081\pm0.0011$&$ 0.0461\pm0.0095$&$0.0000\pm0.0000$&$0.0433\pm0.0102$&$0.0000\pm0.0000$&$\textbf{0.0412}\pm\textbf{0.0106}$\\
		\bottomrule
	\end{tabular}
\end{table*}
\begin{table}[]\footnotesize
	\renewcommand\arraystretch{1.5}
	\setlength{\abovecaptionskip}{0pt}
	\setlength{\belowcaptionskip}{0pt}
	\centering
	\caption{Specification of the datasets}
	\begin{tabular}{cccc}
		\toprule
		\multirow{2}*{Datasets} & \multirow{2}*{Features} & \multicolumn{2}{c}{Observations}\\
		\cline{3-4}
		&{}&Training&Testing\\
		\midrule
		Servo	&5	&83	&83\\
        Airfoil	&5	&751	&751\\
		Concrete	&9	&515 &515\\
        Housing	&14	&253	&253\\
        Yacht	&6	&154	&154\\
        Wine-red &12 &799 &799\\
       	Slump	&10	&52	&51\\
		\bottomrule
	\end{tabular}
\end{table}
\subsection{ELM Based Regression for Benchmark Datasets}

In the second example, we utilize seven benchmark data sets from UCI machine learning repository \cite{23frank2010uci} to confirm the superior regression performance of the MCC-VC based ELM (ELM-MCC-VC) compared with the MCC based ELM (ELM-RCC) \cite{21xing2013training} and regularized ELM (RELM)(\cite{20huang2012extreme}). The descriptions of the data sets are given in Table II. In the simulation, the training and testing samples from each data set are randomly chosen and the data values are normalized into [0, 1]. The parameters of each algorithm are selected through fivefold cross-validation, except that the kernel bandwidth and center location of MCC-VC are chosen through solving (18). We set the kernel center of MCC-VC to the median value of the error samples, only optimize the kernel width $\sigma$ by solving (18). The finite kernel bandwidth set $\mathcal{S}$ is equally spaced over $[0.1,2.0]$ with step size 0.1. The training and testing RMSEs over 100 runs are presented in Table III. Evidently, The ELM-MCC-VC outperforms the ELM-RCC and RELM for all the data sets. Especially on the Yacht data set, MCC-VC can significantly outperform other methods.

\section{Conclusion}

The kernel function in Correntropy is in general a Gaussian function and the kernel center is always located at zero. In this paper, we extended the correntropy to the case where the center can locate at any position. On this basis, the maximum correntropy criterion with variable center (MCC-VC) was proposed. In addition, we proposed an efficient method to optimize the kernel width and center location in MCC-VC. Regression results with linear in parameters (LIP) models have shown the desirable performance of the new method.

\ifCLASSOPTIONcaptionsoff
  \newpage
\fi

\bibliographystyle{IEEEtran}
\bibliography{MCCC}

\end{document}